\documentclass[twocolumn,a4]{revtex4-1}

\usepackage{amsmath,amsfonts,amssymb}
\usepackage{subfigure}
\usepackage{graphics,graphicx}
\usepackage{bbold}

\makeatletter
\newcommand{\xRightarrow}[2][]{\ext@arrow 0359\Rightarrowfill@{#1}{#2}}
\makeatother

\newcommand\ii{\mathcal{I}}
\newcommand\jj{\mathcal{J}}

\begin{document}

\title{On the training of sparse and dense deep neural networks: less parameters, same performance.}

\author{Lorenzo Chicchi$^{1}$, Lorenzo Giambagli$^{1}$, Lorenzo Buffoni$^{1}$, Timoteo Carletti$^{2}$,  Marco Ciavarella$^{1}$, Duccio Fanelli$^{1}$ \vspace*{.25cm}}
\affiliation{$^1$Dipartimento di Fisica e Astronomia, Universit\'a di Firenze, INFN and CSDC, Via Sansone 1, 50019 Sesto Fiorentino, Firenze, Italy}
\affiliation{$^2$naXys, Namur Institute for Complex Systems, University of Namur, Belgium}

\begin{abstract}
Deep neural networks can be trained in reciprocal space, by acting on the eigenvalues and eigenvectors of suitable transfer operators in direct space. Adjusting the eigenvalues, while freezing the eigenvectors, yields a substantial compression of the parameter space. This latter scales by definition with the number of computing neurons.  The classification scores, as measured by the displayed accuracy, are however inferior to those attained when the learning is carried in direct space, for an identical architecture and by employing the full set of trainable parameters (with a quadratic dependence on the size of neighbor layers).  In this Letter, we propose a variant of the spectral learning method as appeared in Giambagli et al {Nat. Comm.} 2021, which leverages on two sets of eigenvalues, for each mapping between adjacent layers. The eigenvalues act as veritable knobs which can be freely tuned so as to (i) enhance, or alternatively silence, the contribution of the input nodes, (ii) modulate the excitability of the receiving nodes with a mechanism which we interpret as the artificial analogue of the homeostatic plasticity. The number of trainable parameters is still a linear function of the network size, but the performances of the trained device gets much closer to those obtained via conventional algorithms, these latter requiring however a considerably heavier computational cost. The residual gap between conventional and spectral trainings can be eventually filled by  employing  a suitable decomposition for the non trivial block of the eigenvectors matrix.  Each spectral parameter reflects back on the whole set of inter-nodes weights, an attribute which we shall effectively exploit to yield sparse networks with stunning classification abilities, as compared to their homologues trained with conventional means.
\end{abstract}

\maketitle

Automated learning from data via deep neural networks \cite{bishop_pattern_2011,cover_elements_1991,hastie2009elements,hundred} is becoming popular in an ever-increasing number of applications \cite{sutton2018reinforcement,graves2013speech,sebe2005machine,grigorescu2020survey}. Systems can learn from data, by identifying distinctive features which form the basis of decision making, with minimal human intervention. The weights, which link adjacent nodes across feedforward architectures, follow the optimisation algorithm and store the information 
needed for the trained network to perform the assigned tasks with unprecedented fidelity \cite{chen2014big, meyers2008using, caponetti2011biologically}. A radically new approach to the training of a deep neural network has been recently proposed which anchors the process to reciprocal space, rather than to the space of the nodes \cite{giambagli2021}.  Reformulating the learning in reciprocal space enables one to shape key collective modes, the eigenvectors, which are implicated in the process of progressive embedding, from the input layer to the detection point. Even more interestingly, one can assume the eigenmodes of the inter-layer transfer operator to align along suitable random directions and identify the  associated eigenvalues as target for the learning scheme. This results in a dramatic compression of the training parameters space, yielding accuracies which are superior to those attained with conventional methods restricted to operate with an identical number of tunable parameters. Nonetheless,  neural networks trained in the space of nodes with no restrictions on the set of adjusted weights,  achieve better classification scores, as compared to their spectral homologues with quenched eigendirections. In the former case, the number of free parameters grows as the product of the sizes of adjacent layer pairs, thus quadratically in terms of hosted neurons. In the latter, the number of free parameters increases linearly with the size of the layers (hence with the number of neurons), when the eigenvalues are solely allowed to change. Also training the eigenvectors amounts to dealing with a set of free parameters equivalent to that employed when the learning is carried out in direct space: in this case, the two methods yield  performances which are therefore comparable.

Starting from this setting, we begin by discussing a straightforward generalisation of the spectral learning scheme presented in \cite{giambagli2021}, which proves however effective in securing a significant
improvement on the recorded classification scores, while still optimising a number of parameters which scales linearly with the size of the network. The proposed generalisation paves the way to a biomimetic interpretation of the spectral training scheme. The eigenvalues can be tuned so as to magnify/damp the contribution associated to the input nodes. At the same time, they modulate the excitability of the receiving nodes, as a function of the local field. This latter effect is reminiscent of the homeostatic plasticity \cite{Surmeier2004} as displayed by living neurons. Further, we will show that the residual gap between conventional and spectral trainings methods can be eventually filled by resorting to apt decompositions of the non trivial block of the eigenvectors matrix, which place the emphasis on a limited set of collective variables. Finally, we will prove that working in reciprocal space turns out to be by far more performant, when aiming at training sparse neural networks. Because of the improvement in terms of computational load, and due to the advantage of operating with collective target variables as we will make clear in the following, it is surmised that modified spectral learning of the type here discussed should be considered as a viable standard for deep neural networks training in artificial intelligence applications. 

To test the effectiveness of the proposed method we will consider classification tasks operated on three distinct database of images. The first is the celebrated MNIST database  of handwritten digits \cite{lecun1998mnist}, the second is Fashion-MNIST (F-MNIST), a dataset of Zalando's article images, the third is CIFAR-10 a collection of images  from different classes (airplanes, cars, birds, cats, deer, dogs, frogs, horses, ships, and trucks). In all considered cases,  use can be made of a deep neural network  to perform the sought classification, namely to automatically assign the image supplied as an input to the class it belongs to. The neural network is customarily trained via standard backpropagation algorithms to tune the weights that connect consecutive stacks of the multi-layered architecture. The assigned weights,  target of the optimisation procedure, bear the information needed to allocate the examined images to their reference category. 

Consider a deep feedforward network made of $\ell$ distinct layers and label each layer with the progressive index $i$ $(=1,...,\ell)$. Denote by  $N_i$ the number of computing units, the neurons, that belong to layer $i$. The total number of parameters that one seeks to optimise in a dense neural network setting (all neurons of any given layer with $i < \ell-1$ are linked to every neurons of the adjacent layer) equals $\sum_{i=1}^{\ell-1} N_i N_{i+1}$, when omitting additional bias. As we shall prove in the following, impressive performance can be also achieved by pursuing a markedly different procedure,  which requires acting on just $N_1+N_{\ell} + 2 \sum_{i=2}^{\ell-1}N_i$ free parameters (not including bias). To this end, let us begin by  reviewing the essence of spectral learning method as set forth in \cite{giambagli2021}. 

Introduce $N=\sum_{i=1}^{\ell} N_i$ and create a column vector $\vec{n}_1$, of size $N$, whose first $N_1$ entries are the intensities (from the top-left to the bottom-right, moving horizontally) as displayed on the pixels of the input image. All other entries of $\vec{n}_1$ are  set to zero. The ultimate goal is to transform $\vec{n}_1$ into an output vector $\vec{n}_{\ell}$ , of  size $N$,  whose last  $N_{\ell}$ elements reflect the intensities of the output nodes where reading takes eventually place. This is achieved with a nested sequence of linear transformations, as exemplified in the following. Consider the generic vector $\vec{n}_{k}$, with $k=1,..., \ell-1$, as obtained after $k$ compositions of the above procedure.  This latter vector undergoes a linear transformation to yield $\vec{n}_{k+1}= {\mathbf A}^{(k)} \vec{n}_{k}$. Further,  $\vec{n}_{k+1}$ is  processed via a suitably defined non-linear function, denoted by $f \left( \cdot, \beta_k \right)$, where $\beta_k$ stands for an optional bias.  Focus now on ${\mathbf A}^{(k)}$, a $N \times N$ matrix with a rather specific structure, as we will highlight hereafter.  Posit ${\mathbf A}^{(k)}={\mathbf \Phi}^{(k)} {\mathbf \Lambda}^{(k)} \left({\mathbf \Phi}^{(k)}\right)^{-1}$, by invoking a spectral decomposition. ${\mathbf \Lambda}^{(k)}$ is the diagonal matrix of the eigenvalues of ${\mathbf A}^{(k)}$. By construction we impose, $\left({\mathbf \Lambda}^{(k)} \right)_{jj} = 1$ for $j< \sum_{i=1}^k N_i$ and $j> \sum_{i=1}^{k+1} N_i$. The remaining $N_k$ elements are initially set to random numbers, e.g. extracted from a uniform distribution, and define the target of the learning scheme \footnote{The only noticeable exception is when $k=1$. In this case, the first $N_1$ diagonal elements of ${\mathbf \Lambda}^{(1)}$ take part to the training.}. Back to the spectral decomposition of  ${\mathbf A}^{(k)}$, ${\mathbf \Phi}^{(k)}$ is assumed to be the identity matrix  $\mathbb{1}_{N \times N}$, with the inclusion of a sub-diagonal rectangular block {\boldmath$\phi$}$^{(k)}$  of size $N_{k+1} \times N_{k}$. This choice corresponds to dealing with a feedforward arrangements of nested layers.  A straightforward calculation returns $\left({\mathbf \Phi}^{(k)}\right)^{-1}=2 \mathbb{1}_{N \times N}- {\mathbf \Phi}^{(k)}$, which readily yields ${\mathbf A}^{(k)}={\mathbf \Phi}^{(k)} {\mathbf \Lambda}^{(k)} \left(2 \mathbb{1}_{N \times N}- {\mathbf \Phi}^{(k)} \right)$. In the simplest setting that we shall inspect in the following,  the off-diagonal elements of matrix ${\mathbf \Phi}^{(k)}$ are frozen to nominal values, selected at random from a given distribution.  In this minimal version, the spectral decomposition of the transfer operators ${\mathbf A}^{(k)}$ enables one to isolate a total of $N=\sum_{i=1}^{\ell} N_i$ adjustable parameters, the full collection of non trivial eigenvalues, which can be self-consistently trained. To implement the learning scheme on these premises, we consider $\vec{n}_{\ell}$, the image on the output layer of the input vector $\vec{n}_{1}$:

\begin{equation}
\label{image}
\vec{n}_{\ell} = f\left(\mathbf{A}^{(\ell-1)}... f\left (\mathbf{A}^{(2)}  f \left (\mathbf{A}^{(1)} \vec{n_1},\beta_1 \right), \beta_2  \right),\beta_{\ell-1} \right) 	  
\end{equation}

and calculate $\vec{z} = \sigma(\vec{n}_{\ell})$ where $\sigma(\cdot)$ stands for the softmax operation. We then introduce the categorical cross-entropy loss function $\text{CE}(l(\vec{n}_1), \vec{z})$ where
the quantity  $l(\vec{n}_1)$ identifies the label attached to $\vec{n}_1$ reflecting the category to which it belongs via one-hot encoding \cite{aggarwal2018NNdeeplearn}. More specifically, the $k$-th elements of vector $l(\vec{n}_1)$ is equal to unit (the other entries being identically equal to zero) if  the image supplied as an input is associated to the class of items grouped under label $k$. 

The loss function can be minimized by acting on a limited set of free parameters, the collection of $N$ non trivial eigenvalues of matrices $\mathbf{A}_{k}$ (i.e. $N_1+N_2$ eigenvalues of $\mathbf{A}^{(1)}$, $N_3$ eigenvalues of $\mathbf{A}^{(2)}$,..., $N_{\ell}$ eigenvalues of $\mathbf{A}^{(\ell-1)}$). In principle, the  sub-diagonal blocks  {\boldmath$\phi$}$^{(k)}$  (the non orthogonal entries of the basis that diagonalises $\mathbf{A}^{(k)}$) can be optimised in parallel, but this choice nullifies the gain in terms of parameters containment, as achieved via spectral decomposition, when the eigenvalues get solely modulated.  The remaining part of this Letter is entirely devoted to overcoming this limitation, while securing the decisive enhancement of the neural network's performance.

The first idea, as effective as it is simple, is to extend the set of trainable eigenvalues. When mapping layer $k$ into layer $k+1$, we can in principle act on $N_{k}+N_{k+1}$  eigenvalues, without restricting the training to the $N_{k+1}$ elements, which were identified as the sole target of the spectral method in its original conception (except for the first mapping, from the input layer to its adjacent counterpart). As we shall clarify in the following, the eigenvalues can be trained twice, depending on whether they originate from incoming or outcoming nodes, along the successive arrangement of nested layers. The global  number of trainable parameters is hence $N_1+N_{\ell} + 2 \sum_{i=2}^{\ell-1}N_i$, as anticipated above. A straightforward calculation, carried out in the annexed Supplementary Information, returns a closed analytical expression for $w^{(k)}_{ij}$, the weights of the edges linking nodes $i$ and $j$ in direct space, as a function of the underlying spectral quantities. In formulae: 

\begin{equation}
\label{w}
w^{(k)}_{ij} = \left( \lambda^{(k)}_{m(j)}-\lambda^{(k)}_{l(i)} \right)  {\Phi}^{(k)}_{l(i), m(j)}
\end{equation}

where $l(i)=\sum_{s=1}^{k} N_s +i$ and $m(j)=\sum_{s=1}^{k-1} N_s +j$, with $i \in \left(1, ..., N_{k+1} \right)$ and $j \in \left(1, ..., N_k \right)$.
More specifically, $j$ runs on the nodes at the departure layer ($k$), whereas $i$ identifies those sitting at destination (layer $k+1$).
In the above expression, $\lambda^{(k)}_{m(j)}$ stand for the first $N_k$  eigenvalues of ${\mathbf \Lambda}^{(k)}$. The remaining $N_{k+1}$ eigenvalues are labelled $\lambda^{(k)}_{l(i)}$.
To help comprehension denote by  $x^{(k)}_j$ the activity on nodes $j$. Then, 

\begin{equation}
x^{(k+1)}_i = \sum_{j=1}^{N_k} \left( \lambda^{(k)}_{m(j)}   {\Phi}^{(k)}_{l(i), m(j)} x_j^{(k)} \right) - \lambda^{(k)}_{l(i)} \sum_{j=1}^{N_k} \left( {\Phi}^{(k)}_{l(i), m(j)} x_j^{(k)} \right).
\end{equation} 

The eigenvalues $\lambda^{(k)}_{m(j)}$ modulate the density at the origin, while $\lambda^{(k)}_{l(i)}$ appears to regulate the local node's excitability relative to the network activity in its 
neighbourhood. This is the artificial analogue of the {\it homeostatic plasticity}, the strategy implemented by living neurons to maintain the synaptic basis for learning, respiration, and locomotion \cite{Surmeier2004}. 

\begin{figure}
\centering
\includegraphics[scale=0.35]{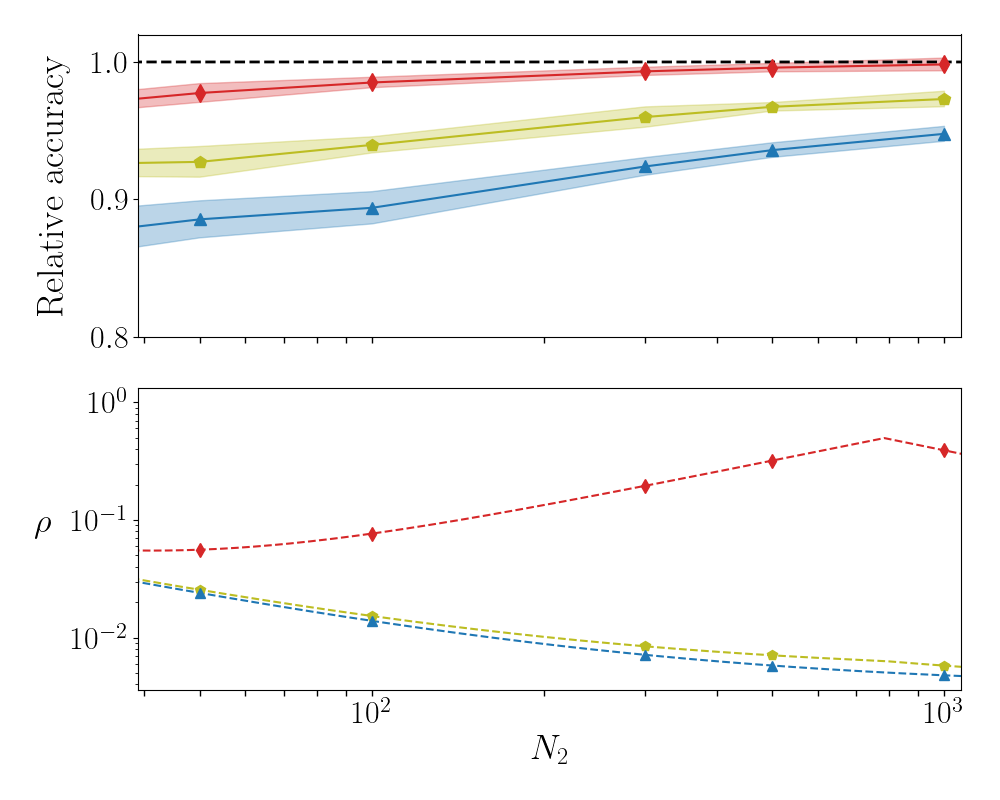} \\ 
\caption{{\bf The case of MNIST.} { Upper panel: the accuracy of the different learning strategies, normalised to the accuracy obtained for an identical deep neural network trained in direct space, 
as a function of the size of the intermediate layer, $N_2$.  Triangles stand for the the relative accuracy obtained when employing the spectral method ({\it Spectral}). Pentagons refer to the setting which extends the training to the eigenvectors' blocks via a SVD decomposition. Specifically, matrices  ${\mathbf U}_k$ and ${\mathbf V}_k$ are randomly generated (with a uniform distribution of the entries) and stay unchanged during  optimisation. The singular values are instead adjusted together with the eigenvalues which stem from the spectral method (this configuration is labelled {\it S-SVD}). Diamonds are instead obtained when  the eigenvalues and the elements of the triangular matrix ${\mathbf R}$ (as follows a QR decomposition of the eigenvectors' blocks)  are simultaneously adjusted {\it S-QR}). Here, ${\mathbf Q}$ is not taking part to the optimisation process (its entries are random number extracted from a uniform distribution). Errors are computed after $10$ independent realisations of the respective procedures. Lower panel: $\rho$ the ratio of the number of tuned parameters (modified spectral, {\it S-SVD}, {\it S-QR} methods vs. conventional one) is plotted against $N_2$.  In calculating $\rho$ the contribution of the bias is properly acknowledged. As a reference, the best accuracy obtained over the explored range for the deep network trained with conventional means is $98 \%$.}}
\label{fig1} 
\end{figure}

To illustrate the effectiveness of the proposed methodology we make reference to Fig. \ref{fig1}, which summarises a first set of results obtained for MNIST. To keep the analysis as simple as possible we have here chosen to deal with $\ell=3$. The sizes of the input ($N_1$) and output ($N_3$) layers are set by the specificity of the considered dataset. Conversely, the size of the intermediate layer ($N_2$) can be changed at will. We then monitor the relative accuracy, i.e. the accuracy displayed by the deep neural networks trained according to different strategies, normalised to the accuracy achieved with an identical network trained with conventional methods.  In the upper panel of Fig. \ref{fig1}, the performance of  the neural networks trained via the modified spectral strategy (referred to as to {\it Spectral})  is displayed in blue (triangles). The recorded accuracy is satisfactory (about 90\% of that obtained with usual means and few percent more than that obtained with the spectral method of original conception \cite{giambagli2021}), despite the modest number of trained parameters. To exemplify this, in the bottom panel of Fig. \ref{fig1} we plot the relative ratio of the number of tuned parameters ({\it Spectral})  vs. conventional one) against $N_2$ (blue triangles) : the reduction in the number of parameters as follows the modified spectral method is staggering. Working with the other employed dataset, respectively F-MNIST and CIFAR-10, yields analogous conclusions (see annexed Supplementary Information).

One further improvement can be achieved by replacing {\boldmath$\phi$}$^{(k)}$ with its equivalent singular value decomposition (SVD), a factorization that generalizes the eigendecomposition to rectangular (in this framework, $N_{k+1} \times N_k$) matrices (see \cite{Gabri__2019} for an application to neural networks). In formulae, this amounts to postulate {\boldmath $\phi$}$^{(k)}$ $= {\mathbf U}_k \mathbf{\Sigma}_k {\mathbf V}_k^T$ where ${\mathbf V}_k$  and ${\mathbf U}_k$ are, respectively,  $N_{k} \times N_{k}$ and $N_{k+1} \times N_{k+1}$ real orthogonal matrices. On the contrary, $\mathbf{\Sigma}_k$ is a $N_{k+1} \times N_k$  rectangular diagonal matrix, with non-negative real numbers on the diagonal. The diagonal entries of $\mathbf{\Sigma}_k$ are the singular values of {\boldmath$\phi$}$_k$.   The symbol $\left( \cdot \right)^T$, stands for the transpose operation.  The learning scheme can be hence reformulated as follows. For each $k$, generate two orthogonal random matrices ${\mathbf U}_k$ and ${\mathbf V}_k$. These latter are not updated during the successive stages of the learning process. At variance, the $M_{k+1}=\min(N_k,N_{k+1})$ non trivial elements of $\mathbf{\Sigma}_k$ take active part to the optimisation process. For each $k$,  $M_{k+1}+N_{k}+N_{k+1}$ parameters can be thus modulated to optimize the information transfer, from layer $k$ to layer $k+1$. Stated differently,  $M_{k+1}$ free parameters adds up to the $N_{k}+N_{k+1}$ eigenvalues that get modulated under the original spectral approach. One can hence count on a larger set of parameters as compared to that made available via the spectral method, restricted to operate with the eigenvalues. Nonetheless, the total number of parameters scales still with the linear size $N$ of the deep neural network, and not quadratically, as for a standard training carried out in direct space. This addition (referred to as the {\it S-SVD} scheme) yields an increase of the recorded classification score, as compared to the setting  where the  {\it Spectral} method is solely employed, which is however not sufficient to fill the gap with conventional schemes (see Fig. \ref{fig1}). Similar scenarios are found for F-MNIST and CIFAR-10 (see Supplementary Information), with varying degree of improvement, which reflects the specificity of the considered dataset.

A decisive leap forward is however accomplished by employing a QR factorization of matrix {\boldmath$\phi$}$^{(k)}$. For $N_{k+1} > N_k$,  this corresponds to writing  the $N_{k+1} \times N_k$ matrix {\boldmath$\phi$}$_k$ as the product of an orthogonal  $N_{k+1} \times N_k$ matrix ${\mathbf Q}_k$ and an upper triangular $N_{k} \times N_k$ matrix ${\mathbf R}_k$.  Conversely, when $N_{k+1} < N_k$, we  factorize {\boldmath$\phi$}$_k^T$, in such a way that the square matrix ${\mathbf R}_k$ has linear dimension $N_{k+1}$.  In both cases, matrix ${\mathbf Q}_k$ is randomly generated and stays frozen during gradient descent optimisation. The $M_{k+1} (M_{k+1}+1)/2$ entries of the $M_{k+1} \times M_{k+1}$ matrix ${\mathbf R}_k$ can be adjusted so as to improve the classification ability of the trained network (this strategy of training, integrated to the {\it Spectral} method, is termed {\it S-QR} ). Results are depicted in  Fig. \ref{fig1} with (red) diamonds. The achieved performance is practically equivalent to that obtained with a conventional approach to learning. Also in this case $\rho<1$, the gain in parameter reduction being noticeable when $N_1$ is substantially different  (smaller or larger) than $N_2$, for the case at hand. Interestingly enough, for a chief improvement of the performance, over the SVD reference case, it is sufficient to train a portion of the off diagonal elements of ${\mathbf R}$. In the Supplementary Information, we report the recorded accuracy against $p$, the probability to train the entries that populate the non null triangular part of ${\mathbf R}_k$. The value of the accuracy attained with conventional strategies to the training is indeed approached, already at values of $p$ which are significantly different from unit.

\begin{figure}
\centering
\includegraphics[scale=0.35]{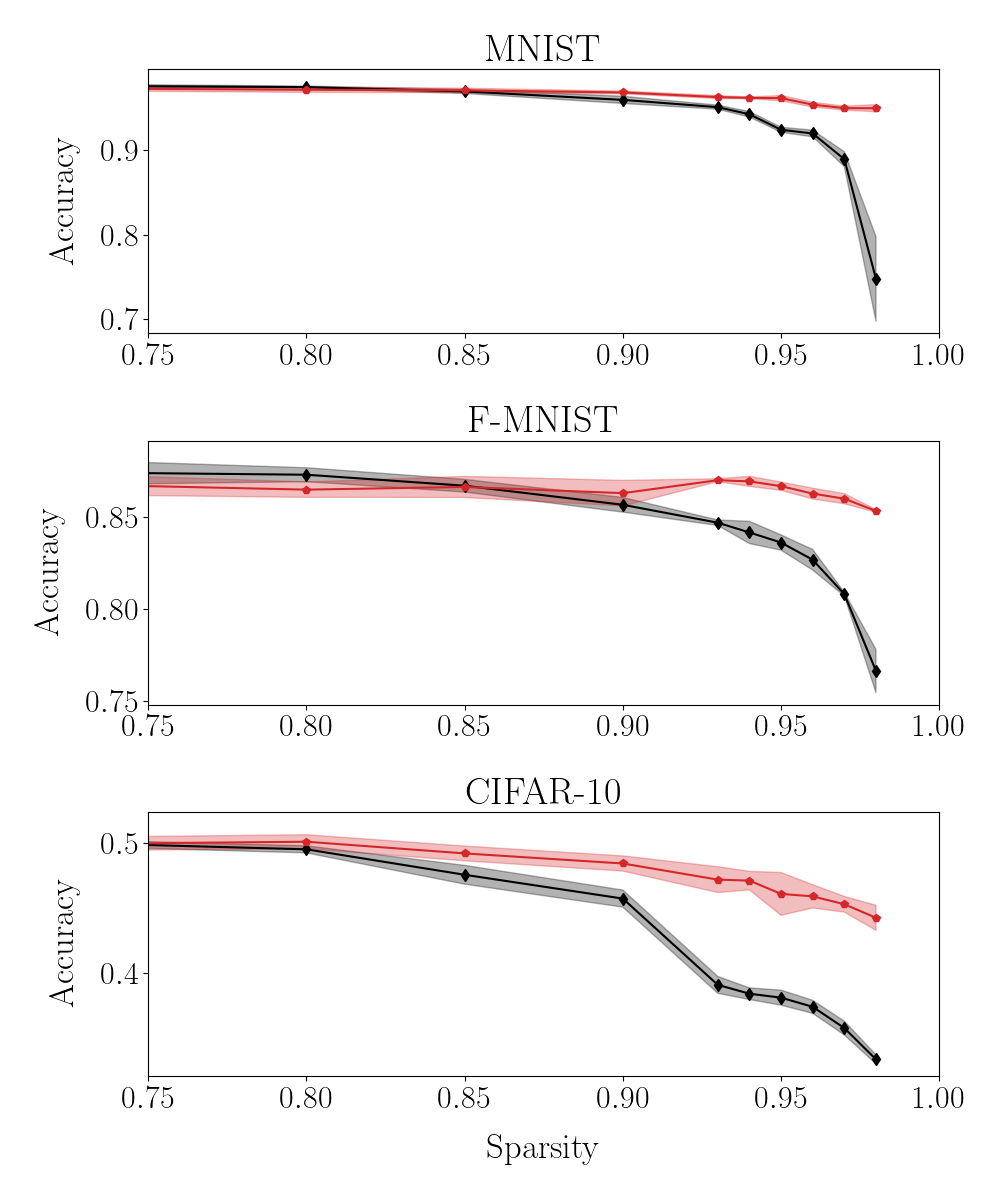} \\ 
\caption{{\bf Training sparse networks.} { The accuracy of the trained network against the degree of imposed sparsity.  Black diamonds refer to the  usual training in direct space, while red pentagons refer to the {\it S-QR} method. From top to bottom: results are reported for MNIST, F-MNIST and CIFAR-10, respectively. In all cases, $\ell=3$. }}
\label{fig2} 
\end{figure}

The quest for a limited subset of key parameters which define the target of a global approach to the training is also important for its indirect implications, besides the obvious reduction in terms of algorithmic complexity. As a key application to exemplify this point, we shall consider the problem of performing the classification tasks considered above, by training a neural network with a prescribed degree of imposed sparsity. This can be achieved by applying a non linear filter on each individual weight $w_{ij}$. The non linear mask is devised so as  to return zero (no link present) when $|w_{ij}|< C$. Here, $C$ is an adaptive cut-off which can be freely adjusted   to allow for the trained network to match the requested amount of sparsity. This latter is measured by a  scalar quantity, spanning the interval $[0,1]$: when the degree of sparsity is set to zero, the network is dense. At the opposite limit, when the sparsity equals one, the nodes of the network are uncoupled and the information cannot be transported  across layers. Working with the usual approach to the training, which seeks to modulate individual weights in direct space, one has to face an obvious problem. When the weight of a given  link  is turned into zero, then it gets excluded by the subsequent stages of  the optimisation process. Consequently, a weight that has been silenced cannot regain an active role in the classification handling. This is not the case when operating under the spectral approach to learning, also when complemented by the supplemental features tested above. The target of the optimisation, the spectral attributes of the transfer operators, are not biased by any filtering masks: as a consequence, acting on them, one can rescue from oblivion weights that are deemed useless at a given iteration (and, as such, silenced),  but which might prove of help, at later stages of the training. In Fig. \ref{fig2}, the effect of the imposed sparsity on the classification accuracy is represented for conventional vs. {\it S-QR} method. The latter is definitely more performant in terms of displayed accuracy, when the degree of sparsity gets more pronounced. The drop in accuracy as exhibited by the sparse network trained with the {\it S-QR} modality is clearly less pronounced, than that reported for an equivalent network optimised in direct space.  Deviations between the two proposed methodologies become indeed appreciable in the very sparse limit, i.e. when the residual active links are too few for a proper functioning of the direct scheme. In fact, edges which could prove central to the classification, but that are set silent at the beginning,  cannot come back to active.  At variance, the method anchored to reciprocal space can identify an optimal pool of links (still constraint to the total allowed for) reversing to the active state, those that were initially set to null. Interestingly, it can be shown that a few hubs emerge in the intermediate layer, which collect and process the information delivered from the input stack.

Taken altogether, it should be unequivocally concluded that a large body of free parameters, usually trained in machine learning applications, is {\it de facto} unessential. The spectral learning scheme, supplemented with a QR training of the non trivial portion of eigenvectors' matrix, enabled us to identify a limited subset of key parameters which prove central to the learning procedure, and reflect back with a global impact on the computed weights in direct space. This observation could materialise in a drastic simplification of current machine learning technologies, a challenge at reach via algorithmic optimisation carried out in dual space. Quite remarkably, working in reciprocal space yields trained networks with better classification scores, when operating at a given degree of imposed sparsity. This finding suggests that shifting the training to the spectral domain might prove beneficial for a wide gallery of deep neural networks applications.

 \bibliographystyle{unsrt}
\bibliography{bibliography}

\begin{thebibliography}{10}

\bibitem{bishop_pattern_2011}
Christopher~M. Bishop.
\newblock {\em Pattern {Recognition} and {Machine} {Learning}}.
\newblock Springer, New York, 1st ed. 2006. corr. 2nd printing 2011 edition
  edition, April 2011.

\bibitem{cover_elements_1991}
T.~M. Cover and Joy~A. Thomas.
\newblock {\em Elements of information theory}.
\newblock Wiley series in telecommunications. Wiley, New York, 1991.

\bibitem{hastie2009elements}
Trevor Hastie, Robert Tibshirani, and Jerome Friedman.
\newblock {\em The elements of statistical learning: data mining, inference,
  and prediction}.
\newblock Springer Science \& Business Media, 2009.

\bibitem{hundred}
\url{http://themlbook.com/}.

\bibitem{sutton2018reinforcement}
Richard~S Sutton and Andrew~G Barto.
\newblock {\em Reinforcement learning: An introduction}.
\newblock MIT press, 2018.

\bibitem{graves2013speech}
Alex Graves, Abdel-rahman Mohamed, and Geoffrey Hinton.
\newblock Speech recognition with deep recurrent neural networks.
\newblock In {\em 2013 IEEE international conference on acoustics, speech and
  signal processing}, pages 6645--6649. IEEE, 2013.

\bibitem{sebe2005machine}
Nicu Sebe, Ira Cohen, Ashutosh Garg, and Thomas~S Huang.
\newblock {\em Machine learning in computer vision}, volume~29.
\newblock Springer Science \& Business Media, 2005.

\bibitem{grigorescu2020survey}
Sorin Grigorescu, Bogdan Trasnea, Tiberiu Cocias, and Gigel Macesanu.
\newblock A survey of deep learning techniques for autonomous driving.
\newblock {\em Journal of Field Robotics}, 37(3):362--386, 2020.

\bibitem{chen2014big}
Min Chen, Shiwen Mao, and Yunhao Liu.
\newblock Big data: A survey.
\newblock {\em Mobile networks and applications}, 19(2):171--209, 2014.

\bibitem{meyers2008using}
Ethan Meyers and Lior Wolf.
\newblock Using biologically inspired features for face processing.
\newblock {\em International Journal of Computer Vision}, 76(1):93--104, 2008.

\bibitem{caponetti2011biologically}
Laura Caponetti, Cosimo~Alessandro Buscicchio, and Giovanna Castellano.
\newblock Biologically inspired emotion recognition from speech.
\newblock {\em EURASIP journal on Advances in Signal Processing}, 2011(1):24,
  2011.

\bibitem{giambagli2021}
Lorenzo Giambagli, Lorenzo Buffoni, Timoteo Carletti, Walter Nocentini, and
  Duccio Fanelli.
\newblock Machine learning in spectral domain.
\newblock {\em Nature Communications}, 12(1):1330, 2021.

\bibitem{Surmeier2004}
Foehring~R. Surmeier, D.
\newblock A mechanism for homeostatic plasticity.
\newblock {\em Nature Neuroscience}, 7, 2004.

\bibitem{lecun1998mnist}
Yann LeCun.
\newblock The mnist database of handwritten digits.
\newblock {\em http://yann. lecun. com/exdb/mnist/}, 1998.

\bibitem{Note1}
The only noticeable exception is when $k=1$. In this case, the first $N_1$
  diagonal elements of ${\protect \mathbf \Lambda }^{(1)}$ take part to the
  training.

\bibitem{aggarwal2018NNdeeplearn}
Charu~C. Aggarwal.
\newblock {\em Neural Networks and Deep Learning}.
\newblock Springer, 2018.

\bibitem{Gabri__2019}
Marylou Gabri{\'{e}}, Andre Manoel, Cl{\'{e}}ment Luneau, Jean Barbier, Nicolas
  Macris, Florent Krzakala, and Lenka Zdeborov{\'{a}}.
\newblock Entropy and mutual information in models of deep neural networks.
\newblock {\em Journal of Statistical Mechanics: Theory and Experiment},
  2019(12):124014, dec 2019.

\end{thebibliography}

\appendix*
 \section{Analytical characterisation of inter-nodes weights in direct space}

In the following, we will derive Eq. (2) as reported in the main body of the paper. Recall that ${\mathbf A}^{(k)}$ is a $N \times N$ matrix. From ${\mathbf A}^{(k)}$ extract a  
square block of size $(N_k+N_{k+1}) \times (N_k+N_{k+1})$. This is formed by the set of elements ${\mathbf A}^{(k)}_{i',j'}$ with $i'=\sum_{s=1}^{k-1}N_s+i$ and $j'=\sum_{s=1}^{k-1}N_s+j$, with $i=1,..., N_k+N_{k+1}$, $j=1,..., N_k+N_{k+1}$. For the sake of simplicity, we use ${\mathbf A}^{(k)}$ to identify the obtained matrix. We proceed in analogy for ${\mathbf \Lambda}^{(k)}$ and ${\mathbf \Phi}^{(k)}$. Then:

\begin{equation}\label{decomp}
		\begin{aligned}
			A_{i j}^{(k)}=&\left[{\mathbf \Phi}^{(k)} {\mathbf \Lambda}^{(k)}\left(2 I-{\mathbf \Phi}^{(k)}\right)\right]_{i j}\\
			=&\left[2 {\mathbf \Phi}^{(k)} {\mathbf \Lambda}^{(k)}\right]_{i j}-\left[{\mathbf \Phi}^{(k)} {\mathbf \Lambda}^{(k)} {\mathbf \Phi}^{(k)}\right]_{i j}\\=& \alpha_{i j}^{(k)}-\beta_{i j}^{(k)}
		\end{aligned}
	\end{equation}                                        

From hereon, we shall omit the apex $(k)$.  Let $ \lambda_1 \dots \lambda_{N_k+N_{k+1}} $ identify the eigenvalues of the transfer operator ${\mathbf A}$, namely 
the diagonal entries of $\Lambda $. In formulae, $ \Lambda _{ij} = \sum_{j = 1}^{N_k+N_{k+1}} \delta_{ij}\lambda_j $.\\
	The quantities $ \alpha_{ij} $ and $ \beta_{ij} $ can be cast in the form:
	\begin{equation*}
		\begin{aligned}
			\alpha_{ij} &=2 \sum_{k=1}^{N_k+N_{k+1}} \Phi_{i k} \lambda_{k} \delta_{k j}=2 \Phi_{i j} \lambda_{j} \\
			\beta_{ij} &=\sum_{k,m = 1}^{N_k+N_{k+1}} \Phi_{i k} \lambda_{k} \delta_{k m} \Phi_{m j}=\sum_{m \in \ii \cup \jj} \delta_{i m} \lambda_{m} \Phi_{m j}
		\end{aligned} 
	\end{equation*}
 where $j \in \mathcal{J}=\left(1, ..., N_k \right)$ runs on the nodes at the departure layer ($k$), whereas $i \in \mathcal{I}=\left(N_k+1,..., N_k+N_{k+1} \right)$. Hence,
$\ii \cup \jj = [1, ..., N_k+N_{k+1}]$. The above expression for $\beta_{ij}$ can be further processed to yield

	\begin{equation*}
		\beta_{i j}=\sum_{m \in J} \Phi_{i m} \lambda_{m} \Phi_{m j}+\sum_{m \in I} \Phi_{i m} \lambda_{m} \Phi_{m j}=\Phi_{i j} \lambda_{j}+\lambda_{i} \Phi_{i j}
	\end{equation*}

	Finally we can express the difference in \eqref{decomp} as
	\begin{equation}\label{pesi}
		\alpha_{i j}-\beta_{i j}=2 \Phi_{i j} \lambda_{j}-\Phi_{i j} \lambda_{j}-\lambda_{i} \Phi_{i j}= (\lambda_j-\lambda_i) \phi_{ij}
	\end{equation}

From the above expression, one readily obtains the sought equation, upon shifting the index $i$ to have it spanning the interval  $[1, ..., N_{k+1}]$. Recall in fact that, by definition, $\mathbf{w}$ (the matrix of the weights, see main body of the paper) is a $N_k \times N_{k+1}$ matrix.

 \section{Testing the  {\it S-SVD} and  {\it S-QR} methods on F-MNIST and CIFAR-10 database}
  
\begin{figure}
\centering
\includegraphics[scale=0.35]{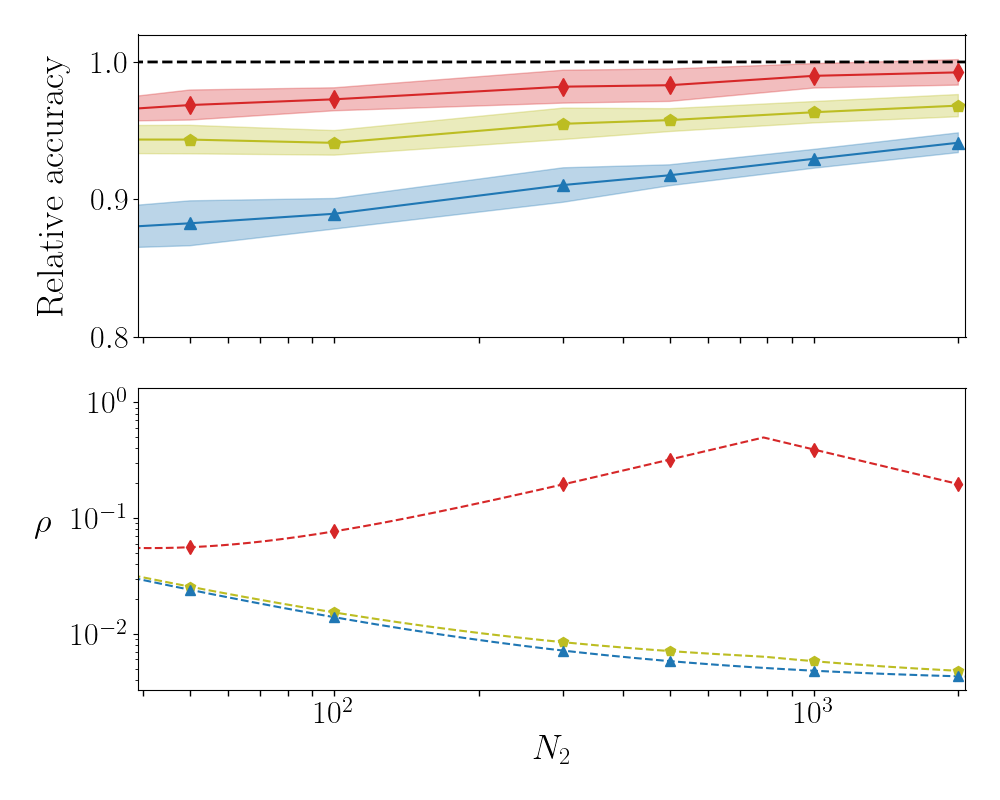} \\ 
\caption{{\bf The case of F-MNIST.} { Upper panel: the accuracy of the different learning strategies, normalised to the accuracy obtained for an identical deep neural network trained in direct space, 
as a function of the size of the intermediate layer, $N_2$.  Triangles stand for the relative accuracy obtained when employing the spectral method ({\it Spectral}). Pentagons refer to the setting which extends the training to the eigenvectors' blocks via a SVD decomposition. Specifically, matrices  ${\mathbf U}_k$ and ${\mathbf V}_k$ are randomly generated (with a uniform distribution of the entries) and stay unchanged during  optimisation. The singular values are instead adjusted together with the eigenvalues which originate from the spectral method ({\it S-SVD}). Diamonds are instead obtained when  the eigenvalues and the elements of matrix ${\mathbf R}$ (in a QR decomposition of the eigenvectors' blocks)  are simultaneously adjusted {\it S-QR}). Errors are computed after $10$ independent realisations of the respective procedures. Lower panel: the ratio of the number of tuned parameters ({\it Spectral}, {\it S-SVD}, {\it S-QR} methods) is plotted against $N_2$.  As a reference, the best accuracy obtained over the explored range for the deep network trained with conventional means is $90 \%$.}}
\label{fig1appB} 
\end{figure}

\begin{figure}
\centering
\includegraphics[scale=0.35]{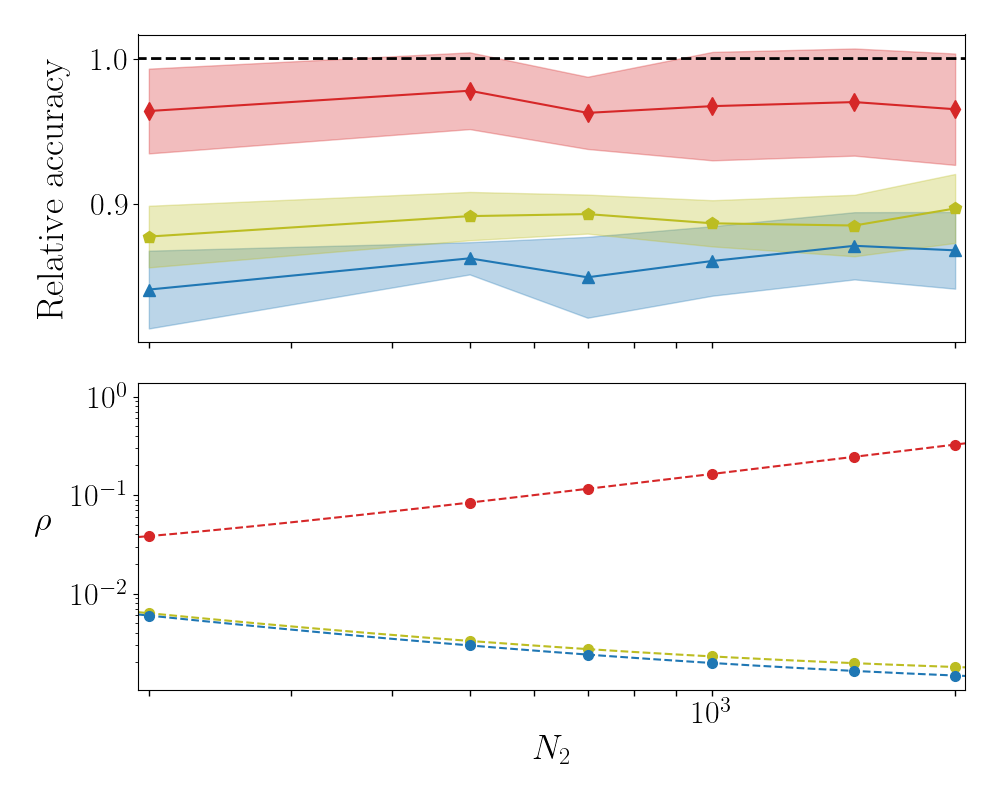} \\ 
\caption{{\bf The case of CIFAR-10.} { Same as in Fig. \ref{fig1appB}. The best accuracy obtained over the explored range for the deep network trained with conventional means is $52 \%$.}}
\label{fig2appB} 
\end{figure}

In the following we report on the accuracy of the {\it S-SVD} and {\it S-QR} methods when applied to the case of F-MNIST and CIFAR-10. The analysis refers to a three layers setting. The results displayed in Figs. \ref{fig1appB} and \ref{fig2appB}  are in line with those discussed in the main body of the paper.
  
\appendix*
 \section{Reducing the number of trainable parameters in the {\it S-QR} method}

\begin{figure}
\centering
\includegraphics[scale=0.35]{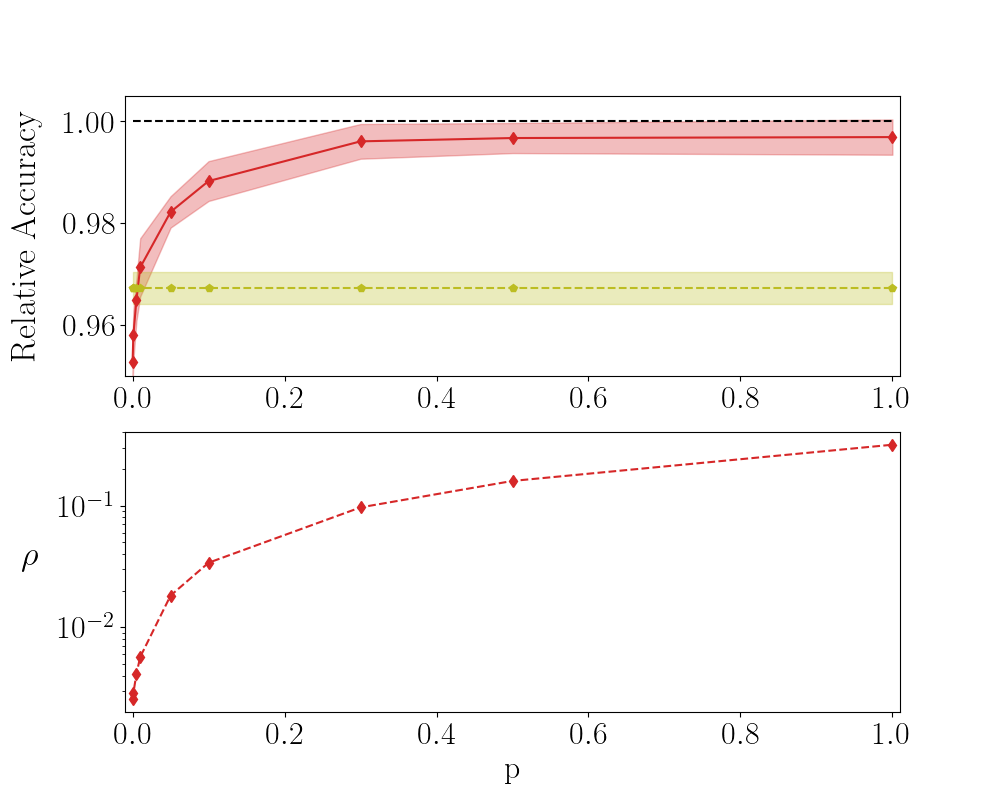} \\ 
\caption{{\bf The case of MNIST.} {The (relative) classification accuracy is plotted (red, diamond and solid line) against $p$, the probability to train the entries that populate the non null triangular part of $R$. The corresponding value of the relative accuracy as computed via the  {\it S-SVD}  is also reported (green, pentagons and solid lines)}. Here, $\ell=3$, with $N_2=500$.}
\label{fig1appC} 
\end{figure}

\begin{figure}
\centering
\includegraphics[scale=0.35]{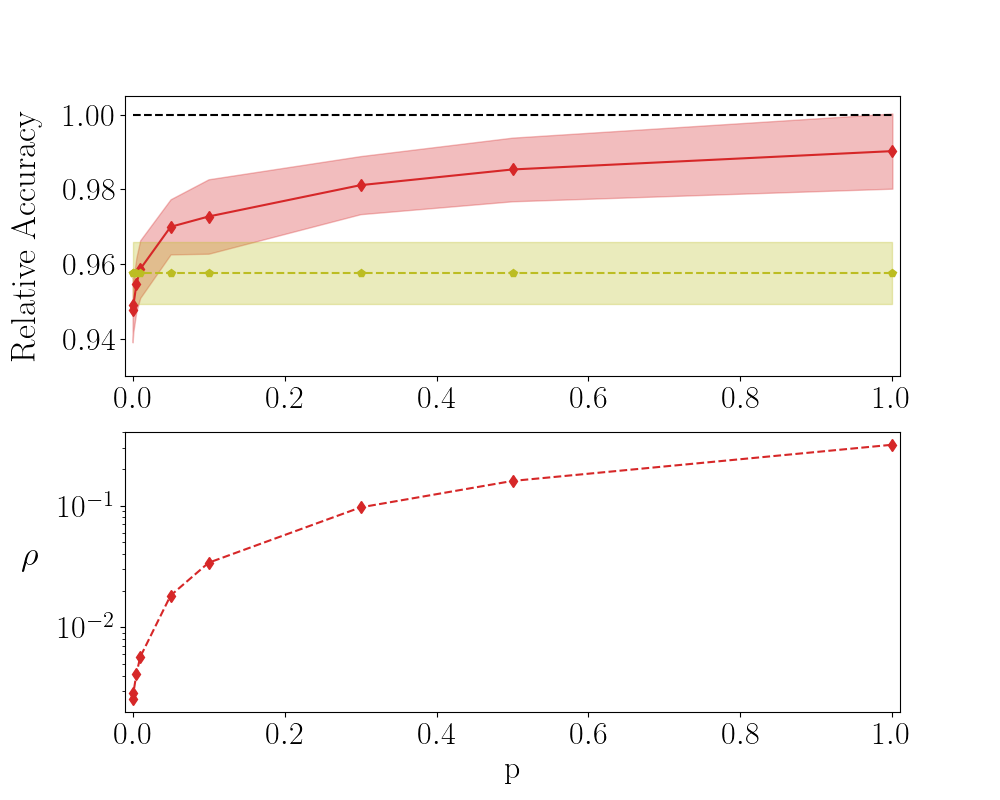} \\ 
\caption{{\bf The case of F-MNIST.} {As in the caption of Fig. \ref{fig1appC}. Here, $N_2=500$.The averages are carried out over 10 independent realisations.}}
\label{fig2appC} 
\end{figure}

\begin{figure}
\centering
\includegraphics[scale=0.35]{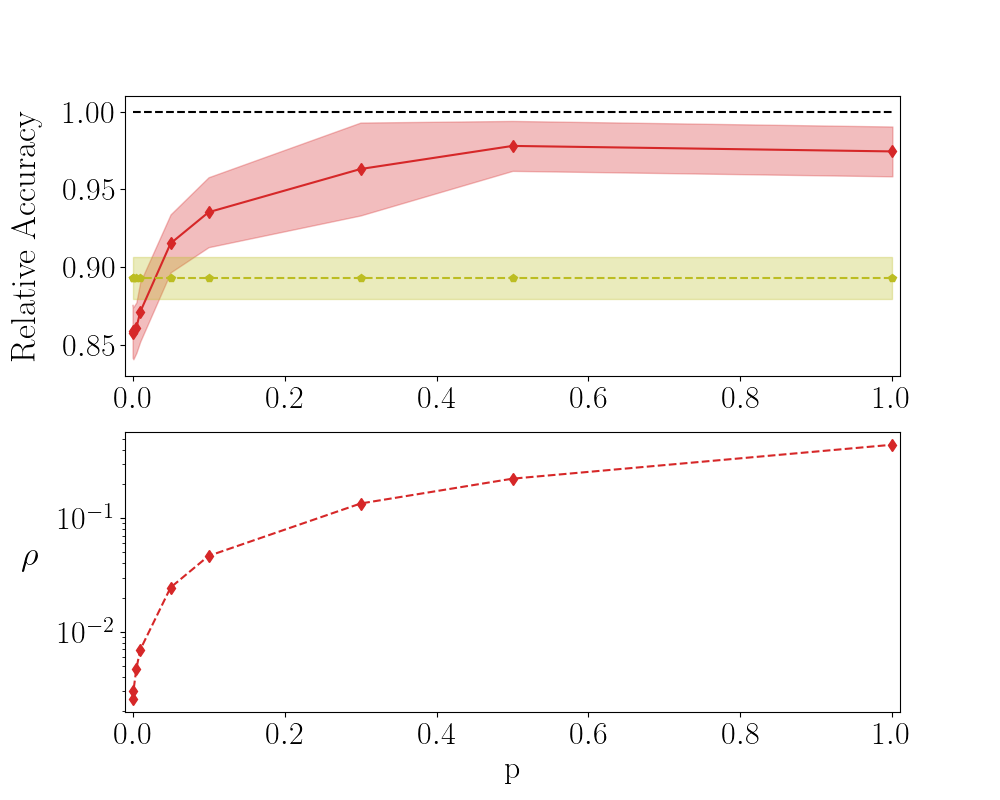} \\ 
\caption{{\bf The case of CIFAR-10.} { As in the caption of Fig. \ref{fig1appC}. Here, $N_2=700$. The averages are carried out over 5 independent realisations.}}
\label{fig3appC} 
\end{figure}

Introduce $p \in [0,1]$. When $p=0$,  the diagonal elements of ${\mathbf R}$ in the  {\it S-QR} method are solely trained. The off-diagonal elements are instead frozen to random values. In the opposite limit, when $p=1$ all elements of matrix ${\mathbf R}$ are assumed to be trained. Intermediate values of $p$ interpolate between the aforementioned limiting conditions. More specifically, the entries that undergo optimisation,  are randomly chosen from the pool of available the available ones, as reflecting the selected fraction. In Fig. \ref{fig1appC} the relative accuracy for MNIST is plotted against $p$. Here, the network is made of $\ell=3$ layers with $N_2=500$. A limited fraction of parameters is sufficient to approach the accuracy displayed by the network trained with conventional means. In Figs. \ref{fig2appC}  and \ref{fig3appC}  the results relative to F-MNIST and CIFAR-10 are respectively reported.

\appendix*
 \section{Testing the performance of the introduced methods on a multi-layered architecture.}

\begin{figure}
\centering
\includegraphics[scale=0.35]{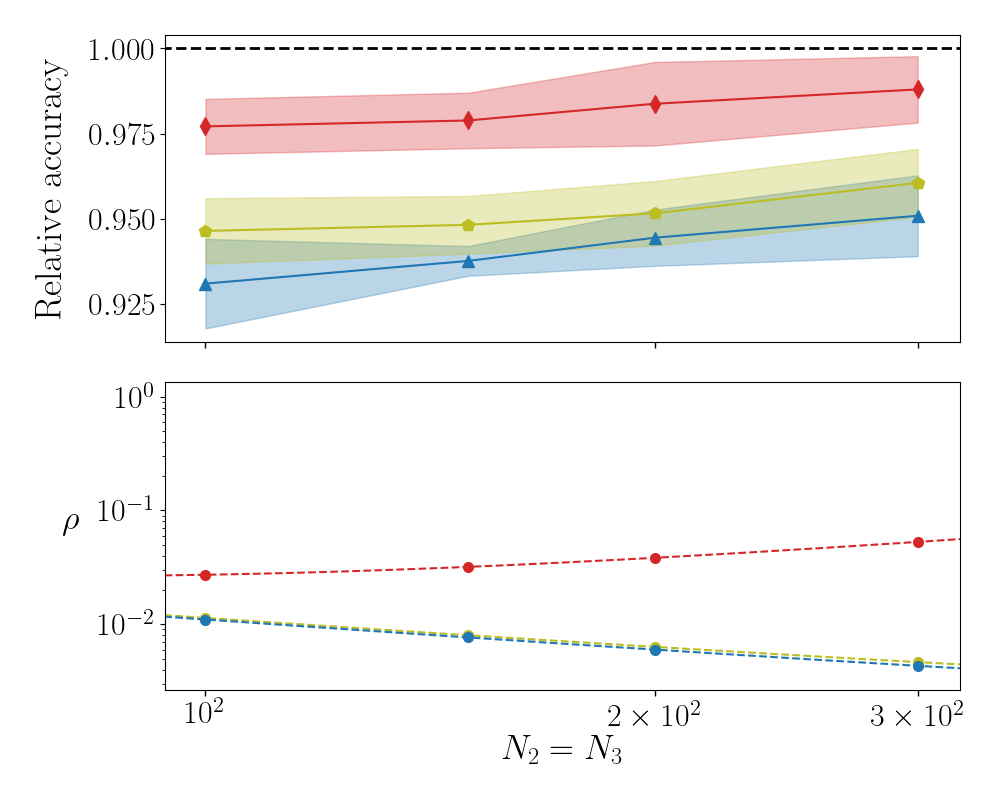} \\ 
\caption{{\bf The case of a multi-layered architecture: the relative accuracy.} {The relative accuracy as obtained by training a four layer network with $N_2=N_3$ via different strategies. The symbols are as specified in Fig. \ref{fig1appB}. The analysis refers to F-MNIST.}}
\label{fig_multi1} 
\end{figure}

\begin{figure}
\centering
\includegraphics[scale=0.35]{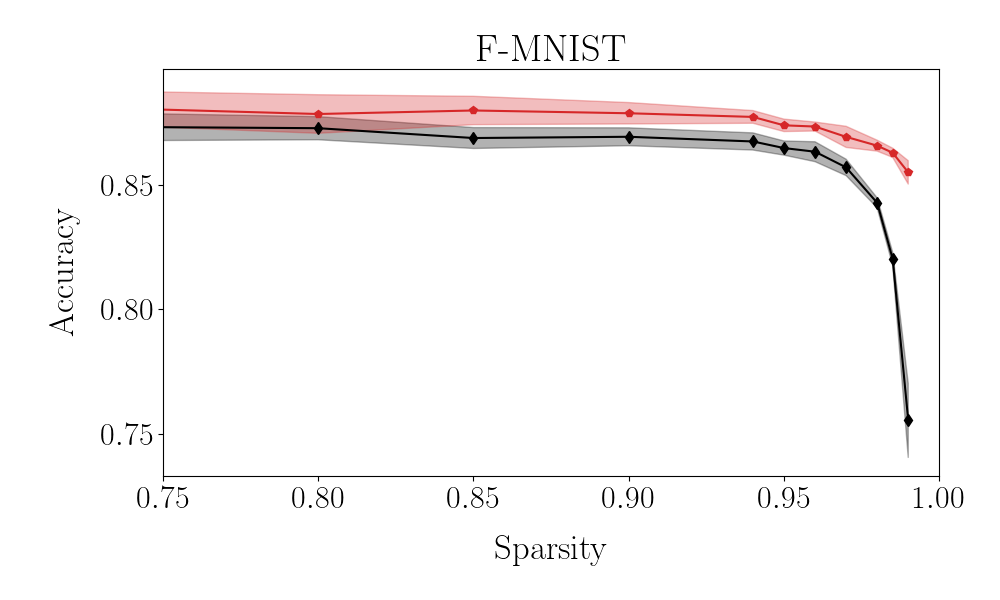} \\ 
\caption{{\bf The case of a multi-layered architecture: training a sparse network.} {The accuracy of the
trained network against the degree of imposed sparsity. Black diamonds refer to the usual training in direct space, while red
pentagons refer to the {\it S-QR} method.The analysis is carried our for F-MNIST. Here, $N_2=N_3=500$.}}
\label{fig_multi2} 
\end{figure}

In this Section we will test the setting of a multi-layered architecture by generalising beyond the case study $\ell=3$ that we employed in the main body of the paper. More specifically, we have trained according to different modalities a four-layer ($\ell=4$) deep neural network, by modulating $N_2=N_3$ over a finite window. As usual, the size of the incoming and outgoing layers are set by the specificity of the examined datasets. The results reported in Fig. \ref{fig_multi1} refer to F-MNIST and confirm that the {\it S-QR}  strategy yields performance that are comparable to those reached with conventional learning approaches, but relying on a much smaller set of trainable parameters. In Fig. \ref{fig_multi2} the effect of the imposed sparsity on the classification accuracy is displayed for both conventional 
and {\it S-QR} method. Similar conclusions can be reached for MNIST and CIFAR-10.
 
 \end{document}